\documentclass{esann}
\usepackage[dvips]{graphicx}

\usepackage{booktabs}           
\usepackage{multirow}           
\usepackage{amsfonts}           
\usepackage{graphicx}           
\usepackage{duckuments}         
\usepackage{float}
\usepackage{algorithm}
\usepackage{amsmath}
\usepackage{array}
\usepackage{tabularx}
\usepackage{booktabs}
\usepackage{multirow}
\usepackage{color,epsfig}
\usepackage{cite}
\usepackage[caption = false,font = footnotesize]{subfig}
\usepackage{url}
\usepackage{a4wide}
\usepackage[framemethod=tikz]{mdframed}
\usepackage{algorithmic}    

\newcommand{\vc}[1]{\boldsymbol{\mathbf{#1}}}

\definecolor{carnelian}{rgb}{0.7, 0.11, 0.11}
\definecolor{ao(english)}{rgb}{0.0, 0.5, 0.0}
\definecolor{bleudefrance}{rgb}{0.19, 0.55, 0.91}
\definecolor{dif}{rgb}{0.0, 0.0, 0.0}

\title{Combining Stochastic Explainers and Subgraph Neural Networks can Increase Expressivity and Interpretability}

\author{Indro Spinelli $^1$ and Michele Guerra $^2$ and Filippo Maria Bianchi $^{2,3}$ and Simone Scardapane $^4$
\thanks{This work was partially supported by the CHIST-ERA grant CHIST-ERA-19-XAI-009 and PNRR MUR project PE0000013-FAIR.}
\vspace{.3cm}\\
1 - Sapienza University of Rome - DI \\
Via Salaria 113, 00198 Rome - Italy \\
\vspace{.1cm}\\
2 - UiT the Arctic University of Norway - Department of Mathematics and Statistics \\
Hansine Hansens veg 18, 9019 Troms\o{} - Norway \\
\vspace{.1cm}\\
3 - NORCE Norwegian Research Centre \\ Nyg\r{a}rdstangen 22, 5838 Bergen - Norway \\
\vspace{.1cm}\\
4 - Sapienza University of Rome - DIET \\
Via Eudossiana 18, 00184 Rome - Italy \\
}

\begin{document}

\maketitle

\begin{abstract}
Subgraph-enhanced graph neural networks (SGNN) can increase the expressive power of the standard message-passing framework. This model family represents each graph as a collection of subgraphs, generally extracted by random sampling or with hand-crafted heuristics. Our key observation is that by selecting ``meaningful'' subgraphs, besides improving the expressivity of a GNN, it is also possible to obtain interpretable results. For this purpose, we introduce a novel framework that jointly predicts the class of the graph and a set of explanatory sparse subgraphs, which can be analyzed to understand the decision process of the classifier. We compare the performance of our framework against standard subgraph extraction policies, like random node/edge deletion strategies. The subgraphs produced by our framework allow to achieve comparable performance in terms of accuracy, with the additional benefit of providing explanations.
\end{abstract}

\section{Introduction}
Graph neural networks (GNNs) are neural network models designed to adapt and perform inference on graph domains \cite{bacciu2020gentle}. While a few models were already proposed in-between 2005 and 2009 \cite{gori2005new,micheli2009neural}, the interest in GNNs has increased dramatically over the last few years, thanks to the broader availability of data, processing power, and automatic differentiation frameworks. Now, GNNs are the state-of-the-art solution in a comprehensive set of scenarios. Nevertheless, regulations require high task performance and a transparent decision process \cite{shen2022trust}.

For this reason, several researchers have investigated techniques to explain GNNs' predictions, primarily identifying the most critical portions of the graph that contributed to producing a particular inference. The vast majority of these techniques \cite{ying2019gnnexplainer,luo2020parameterized} provide a post-hoc explanation, thus inferring the reasons that led to a specific outcome by a trained model. However, recent efforts toward ``explainable-by-design" GNNs rather than post-hoc explainers are opening up new, interesting approaches. For example, in \cite{spinelli2022mate}, the authors introduce an explainability term to let the network converge to a ``interpretable" local minimum to facilitate the work of post-hoc explanation algorithms. Solutions like \cite{zang2021prot,yu2021GIB} completely discard the notion of post-hoc algorithms providing explanations directly in the main model output.

On a separate line of research, recent studies \cite{bevilacqua2021esan,cotta2021, chendi2022ordered} demonstrated that by providing the GNN with subgraphs that give different views of the same graph, it is possible to increase the expressive power of the standard message-passing framework.

We propose to connect these two topics and build an explainable by-design subgraph-enhanced GNN. We use a data-driven approach to learn small and representative subgraphs that increase the expressive power for the downstream task and that can be used as explanations.

\section{Related Works}
\subsection{Explainability in GNNs}
GNNs are generally seen as a black-box since they form decisions in high-dimensional and nonlinear spaces. However, explainability techniques enable humans to interpret the decision process of GNNs, discover potential sources of error, find biases and limitations in the model and learn more about the data and the task at hand.
Most approaches in the literature are post-hoc explainers differentiated according to the techniques used to explain a trained GNN. In particular, there are
gradient-based approaches \cite{pope2019exp}, perturbation-based approaches\cite{ying2019gnnexplainer,luo2020parameterized,schlichtkrull2020gradmask}, decomposition methods \cite{schnake2021},
 and counterfactual explainers like \cite{lucic2021cf}  and GEM \cite{lin2021gem}.
 
Of particular interest for this work is \textbf{PGExplainer} \cite{luo2020parameterized}, which uses a small network to parametrize the probability of each edge $\omega_{ij}$ of being part of the explanatory subgraph, and sample from this distribution to obtain the final explanation subgraph characterized by edges $e_{ij}$. The optimization objective maximizes the mutual information between the masked prediction and the prediction obtained from the original graph. In addition, an element-wise entropy term encourages sparsity on $\omega$, and an $l1-$norm forces small explanation subgraphs.

Fewer research efforts have been devoted to explainable-by-design GNNs. ProtGNN \cite{zang2021prot} first learns some subgraph prototypes from the dataset. Then, they compare the inputs to these prototypes in the latent space. Furthermore, they introduce a subgraph sampling algorithm to highlight which component in the input is most similar to each learned prototype. GIB \cite{yu2021GIB}, instead, uses a bi-level optimization scheme to find the IB-subgraph, which is the most informative yet compressed subgraph. This subgraph is the one that maximizes the mutual information (MI) or shares the same properties with the original input. GSAT \cite{miao2022gsat}, which builds upon GIB, injects stochasticity to the attention weights to block the information from task-irrelevant graph components while learning task-relevant subgraphs for interpretation.

\subsection{Subgraph-enhanced graph neural networks}
The study of the expressive power of GNNs has always been of central interest to the community. 
Most GNNs operating via local neighbourhood aggregation are as powerful as the Weisfeiler–Leman (1-WL) graph-isomorphism test~\cite{xu2018how}.
It has recently been shown that it is possible to create more expressive GNNs using standard architectures that process several subgraphs of the input graph \cite{cotta2021, bevilacqua2021esan}. 
In particular, \textbf{ESAN} \cite{bevilacqua2021esan} represents each graph as a bag of subgraphs $\{\mathcal{G}_1,\dots,\mathcal{G}_m\}$ chosen according to some predefined policy, e.g., all graphs obtainable by removing one edge (edge deleted strategy) or one node (node deleted strategy) from the original graph. The encoder implements a module $L_1$ consisting of several message-passing layer that process each subgraph independently, and then a second message-passing module $L_2$ preprocesses the aggregation of the subgraphs working as an information-sharing module for the subgraphs. For example, to compute the embedding $H_i$ for the subgraph $\mathcal{G}_i$, they use the following procedure:

\begin{equation}
    \text H_i = L_1\left(\mathcal{G}_i \right ) + L_2\left(\sum_j \mathcal{G}_j \right )\,. 
\end{equation}

Then, in the last layer of the encoder, a pooling operation aggregates the node embeddings into subgraph embeddings with a global pooling operation. Finally, a set learning module \cite{maron2020sym} aggregates the obtained subgraph representations into a single one used in downstream tasks. The authors of \cite{bevilacqua2021esan} name this general configuration DSS-GNN. 

Recently k-OSANs \cite{chendi2022ordered} developed a data-driven policy for the subgraph selection that improves the predictive performances compared to the simple policies used in ESAN.

\section{Proposed Framework}

In this work, we consider an undirected and unweighted graph $\mathcal{G} = (\mathcal{V}, \mathcal{E})$, where $\mathcal{V} = \left\{1, \ldots, n\right\}$ is the set of node indexes, and $\mathcal{E} \textcolor{dif}{\subseteq} \left\{(i, j) \; \vert \; i, j \in \mathcal{V}\right\}$ is the set of edges connecting pairs of nodes. The entities and relationships represented by nodes and edges depend on the data and the application. 
The graph topology can be represented by the adjacency matrix $\vc{A} \in \left\{0, 1\right\}^{n\times n}$. Other operators matching the sparsity pattern of $\vc{A}$, such as the graph Laplacian, can be used to define the weighted connectivity of the graph. Letting $\vc{D}$ be one of these operators encoding the topological information, a GNN layer is defined by (ignoring biases for simplicity, but w.l.o.g.):
\begin{equation}
    \mathbf{H} = \phi\left( \mathbf{D}\mathbf{X}\mathbf{W} \right) \,,
    \label{eq:gc_layer}
\end{equation}
where $\mathbf{X} \in \mathbb{R}^{n \times d}$ is a matrix collecting all vertex features row-wise, $\mathbf{W}\in\mathbb{R}^{d \times q}$ is a matrix of trainable coefficients, and $\phi$ is an element-wise non-linear function (e.g., for ReLU $\phi(s)=\max\left(0, s\right)$). A GNN can stack multiple layers in the form of \eqref{eq:gc_layer} to learn highly nonlinear node representations that account for larger neighbourhoods on the graph. In this work, we are interested in graph classification tasks. This task requires a global pooling operation, such as mean, max, or min pooling, to build a graph representation from the node representations produced by stacking the layers described in Equation \eqref{eq:gc_layer}.

\begin{figure*}
    \centering
    \includegraphics[width=0.62\textwidth]{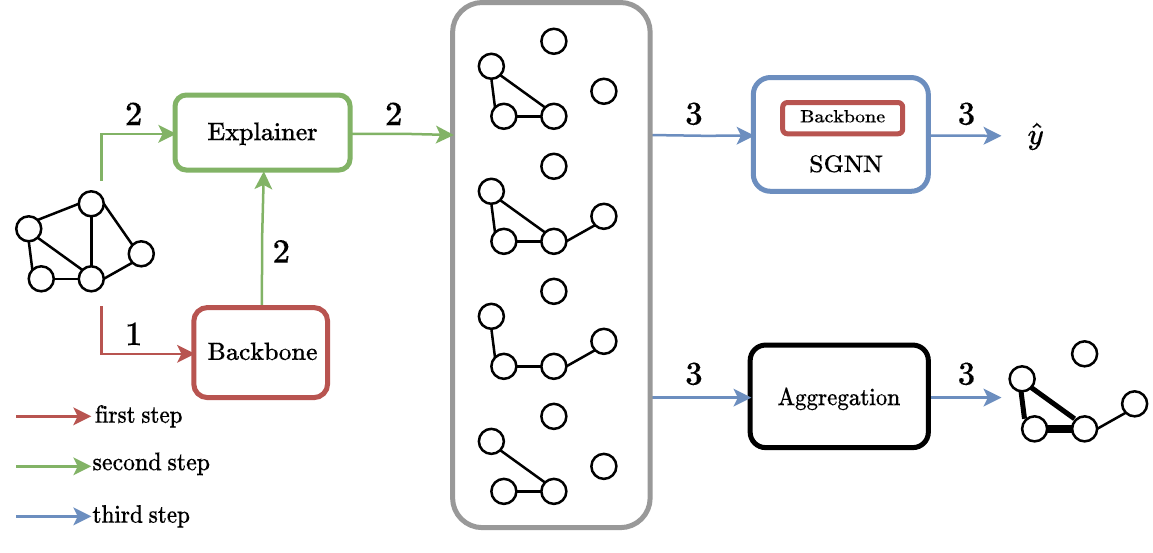}
    \caption{In the \textcolor{carnelian}{first step}, we train the backbone of the SGNN, a GIN classifier, using the original graphs. In the \textcolor{ao(english)}{second step}, we train the explainer with the original graph and the backbone's predictions. We then create the new representation consisting of bags of explanation subgraphs. Finally, in the \textcolor{bleudefrance}{third step}, we train the whole SGNN framework, fine-tuning the backbone using the explanatory subgraphs. Our model outputs the predicted label and an explanation obtained by combining all the subgraphs used during training.}
    \label{img:citeseer}
\end{figure*}

Our goal is to develop a framework that jointly predicts the graph class and the explanation masks, highlighting the parts of the graph that contribute the most to the prediction. The key to our framework is the role played by the subgraphs which are learned end-to-end, based on the optimization of the classification loss. 
Firstly, they must improve the expressive power of the base GNN classifier. 
Secondly, they must serve as explanation masks by selecting the parts of the input that mostly contribute to determine the correct class. 
We expect such subgraphs to be more informative than post-hoc explanation masks since they are directly generated by the model to maximize the classification performance. 

We summarize our framework, which consists of three main steps highlighted in Fig.~\ref{img:citeseer}. 
In the \textcolor{carnelian}{first step}, we train the backbone of the SGNN, which is a classifier that processes the original graph. 
In the \textcolor{ao(english)}{second step} we apply PGExplainer with a minor modification to circumvent the ``introduced evidence" \cite{dabkowski2017soft} issue due to the presence of soft masks. Specifically, we binarize the soft weights to create ``hard'' explanation masks. 
Since ``hard'' explanations are not differentiable, to allow the computation and the flowing of the gradients in the backward pass, we apply a straight-through estimator \cite{hinton2012ste}. These modifications lead to design choices for the regularization terms that are different from those originally used in PGExplainer. In particular, we apply the regularization directly on the ``hard'' explanation mask $e$ instead of $\omega$. Therefore we remove the element-wise entropy term and the $l1-$norm, in this case, is equivalent to an $l0-$norm minimizing the number of edges appearing in the explanation. 

\begin{figure*}
    \centering
    \includegraphics[width=0.40\columnwidth]{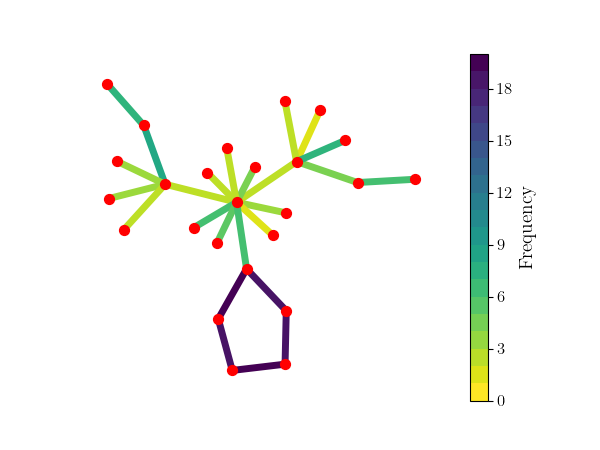}
    \includegraphics[width=0.40\columnwidth]{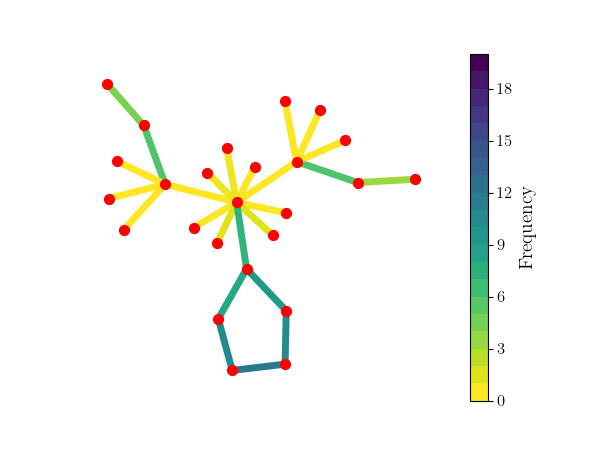}
    \caption{The collection of subgraphs is represented as a heatmap.
    On the left, there are the subgraphs obtained with the noise based strategy. On the right, those obtained with the progressive top-K selection. The motif responsible for the prediction is clearly visible in both cases, however, the top-K approach selects less spurious edges.}
    \label{img:heatmaps}
\end{figure*}

While PGExplainer generates a unique explanation subgraph, we are interested in collecting several subgraphs to train the subgraph-based classifier. Therefore, we devise two strategies to obtain a heterogeneous yet meaningful set of subgraphs. The first is to use the reparametrization trick also in the inference phase allowing the noise, which can be tuned as a hyperparameter, to inject diversity that leads to multiple different subgraphs. In the second approach we introduce a budget on the explanation. We adapt the binarization threshold to obtain a set of explanations with $K$ edges, where $K$ ranges from $5\%$ to $75\%$ of the initial number of edges in the original graph. An example of these two strategies is presented in Figure \ref{img:heatmaps}, on the synthetic dataset BA-2Motifs where the five nodes cycle is responsible for the prediction.

In the \textcolor{bleudefrance}{third step}, we insert the backbone GNN into the DSS-GNN model introduced in ESAN in \cite{bevilacqua2021esan}. The backbone works as the encoder that processes each explanation subgraph independently. A new message-passing module preprocesses the aggregation of the explanations working as an information-sharing module for the subgraphs. Finally, a new set learning module \cite{maron2020sym} aggregates subgraph representations obtained after a global pooling operation into a single one used in downstream tasks.
Besides being used by the information-sharing module, the subgraph obtained by aggregating all the explanations is used to explain the model's prediction. We would like to stress the fact that this explanation, rather than being computed just from a post-hoc explainer, is used to train the model.

\section{Experimental Evaluation}

We want to prove that our framework retains the same classification performance of known SGNN architectures. However, our framework has the added benefit of re-using the subgraphs involved in the computations as plausible explanations for the model's prediction. We used the same evaluation proposed in ESAN \cite{bevilacqua2021esan} over the TUD repository datasets. This consist of computing the average accuracy obtained by running a $10-$fold cross-validation. We selected GIN \cite{xu2018how} and DSS-GNN with the GIN backbone as baselines. We used the most compact model consisting of 4 GIN layers with two linear layers of a hidden size of 32. The batch size is also 32. We selected the top performing policies from \cite{bevilacqua2021esan}: node deleted (ND) and edge deleted (ED).
Furthermore, we consider both their deterministic versions: one that uses the entire bag of subgraphs (referred to as ``1.0'') and the sampled versions, which keeps only $10\%$ of the bag (referred to as ``0.1''). Finally, we select our hyperparameters using grid-search. We report the results in Table \ref{tab:TUD}.
\begin{table}[t]
        \centering
        \[
        \begin{array}{l|ccccc}
            \toprule
            \text{Model} & \text{MUTAG} & \text{NCI1} & \text{PROTEINS} & \text{IMDB-BINARY} & \text{IMDB-MULTI}\\
            \midrule
            \text{GIN} & 89.4 \pm5.6 & 82.7\pm1.7 & 76.2\pm2.8 & 75.1\pm5.1 & 52.3\pm2.8 \\
            \text{DSS ND }1.0 & 89.0\pm4.4 & 83.0\pm2.0 & 76.5\pm3.7 & 75.4\pm3.5 & 52.3\pm2.9\\
            \text{DSS ND } 0.1 & 89.9\pm2.8 & \textbf{83.1}\pm2.0 & \textbf{76.6}\pm4.7 & 75.7\pm2.3 & 52.5\pm2.2 \\
            \text{DSS ED } 1.0 & 89.9\pm4.3 & 82.7\pm1.7 & 74.0\pm4.4 & \textbf{76.7}\pm3.6 & 52.6\pm2.6\\
            \text{DSS ED } 0.1 & 89.9\pm5.5 & 82.1\pm2.0 & 75.1\pm4.7 & 76.4\pm2.7 & 52.6\pm2.8\\
            \text{OURS} & \textbf{92.1} \pm 4.3 & 82.9\pm2.3 & 76.3 \pm 4.4 & 76.0 \pm 3.1 &  \textbf{53.5} \pm 3.8\\
            \bottomrule
        \end{array}
        \]
        \caption{TUDataset classification accuracy. We report the mean and the standard deviation obtained by running a $10-$fold cross-validation.}
        \label{tab:TUD}
    \end{table}
We can observe that subgraph-enhanced GNNs achieve slightly better accuracies. We did not notice any empirical difference when using a subset of the bag of subgraphs instead of the complete one. Furthermore, as noted in \cite{bevilacqua2021esan}, despite the model being no longer invariant, the sample bag of subgraphs still increases the expressive power of the base encoder. Finally, the results obtained using explanatory subgraphs instead of randomly generated ones are comparable. However, we have the advantage that our bag of subgraphs forms an explanation helping human experts to extract new knowledge from the model.

\section{Conclusion}
We introduced a framework to bridge subgraph-enhanced GNNs with explainability techniques.
First, we pre-train a base GNN such that we can train a post-hoc explainer modified to satisfy the need for multiple and discrete explanation masks. Next, the explainer generates a bag of explanation subgraphs. Then, we resume the training of the base model, including equivariant layers, to process the new representation in the SGNN fashion. Finally, we combine the subgraphs in a unique heatmap highlighting the most relevant parts responsible for the prediction and the prediction itself. This additional information allows human experts to interpret and extract knowledge from the model. The experimental evaluation showed that our model achieves comparable classification performance to state-of-the-art models while providing, at the same time, an explanation.

\bibliographystyle{unsrt}
\bibliography{bib}
\end{document}